\documentclass[11pt,a4paper]{article}
\usepackage[hyperref]{naaclhlt2018}
\usepackage{times}
\usepackage{latexsym}
\usepackage{graphicx}
\usepackage{amsmath}
\usepackage{url}
\usepackage{multirow}
\usepackage{enumitem}
\usepackage{CJKutf8}
\usepackage{xcolor}

\usepackage{url}

\aclfinalcopy 
\def\aclpaperid{93} 

\title{Handling Homographs in Neural Machine Translation}

\author{Frederick Liu\thanks{$^*$Equal contribution.} \thanks{$^\dagger$Now at Snap Inc.} , Han Lu\footnotemark[1] \thanks{$^\ddagger$Now at Google} , Graham Neubig \\
  Language Technologies Institute \\
  Carnegie Mellon University \\
  {\tt \{fliu1,hlu2,gneubig\}@cs.cmu.edu}}

\date{}

\begin{document}
\maketitle
\begin{abstract}
Homographs, words with different meanings but the same surface form, have long caused difficulty for machine translation systems, as it is difficult to select the correct translation based on the context. However, with the advent of neural machine translation (NMT) systems, which can theoretically take into account global sentential context, one may hypothesize that this problem has been alleviated. In this paper, we first provide empirical evidence that existing NMT systems in fact still have significant problems in properly translating ambiguous words. We then proceed to describe methods, inspired by the word sense disambiguation literature, that model the context of the input word with context-aware word embeddings that help to differentiate the word sense before feeding it into the encoder. Experiments on three language pairs demonstrate that such models improve the performance of NMT systems both in terms of BLEU score and in the accuracy of translating homographs.%
\footnote{Code for our translation models is available at \\ \url{https://goo.gl/oaiqoT}}
\end{abstract}

\section{Introduction}

Neural machine translation (NMT; \newcite{sutskever2014sequence,bahdanau2015align}, \S\ref{sec:nmt}), a method for MT that performs translation in an end-to-end fashion using neural networks, is quickly becoming the de-facto standard in MT applications due to its impressive empirical results.
One of the drivers behind these results is the ability of NMT to capture long-distance context using recurrent neural networks in both the encoder, which takes the input and turns it into a continuous-space representation, and the decoder, which tracks the target-sentence state, deciding which word to output next.
As a result of this ability to capture long-distance dependencies, NMT has achieved great improvements in a number of areas that have bedeviled traditional methods such as phrase-based MT (PBMT; \newcite{koehn2003statistical}), including agreement and long-distance syntactic dependencies \cite{neubig2015neural,bentivogli2016neuralvsphrasebased}.

\begin{figure}[!t]
\centering
\includegraphics[width= 1\linewidth]{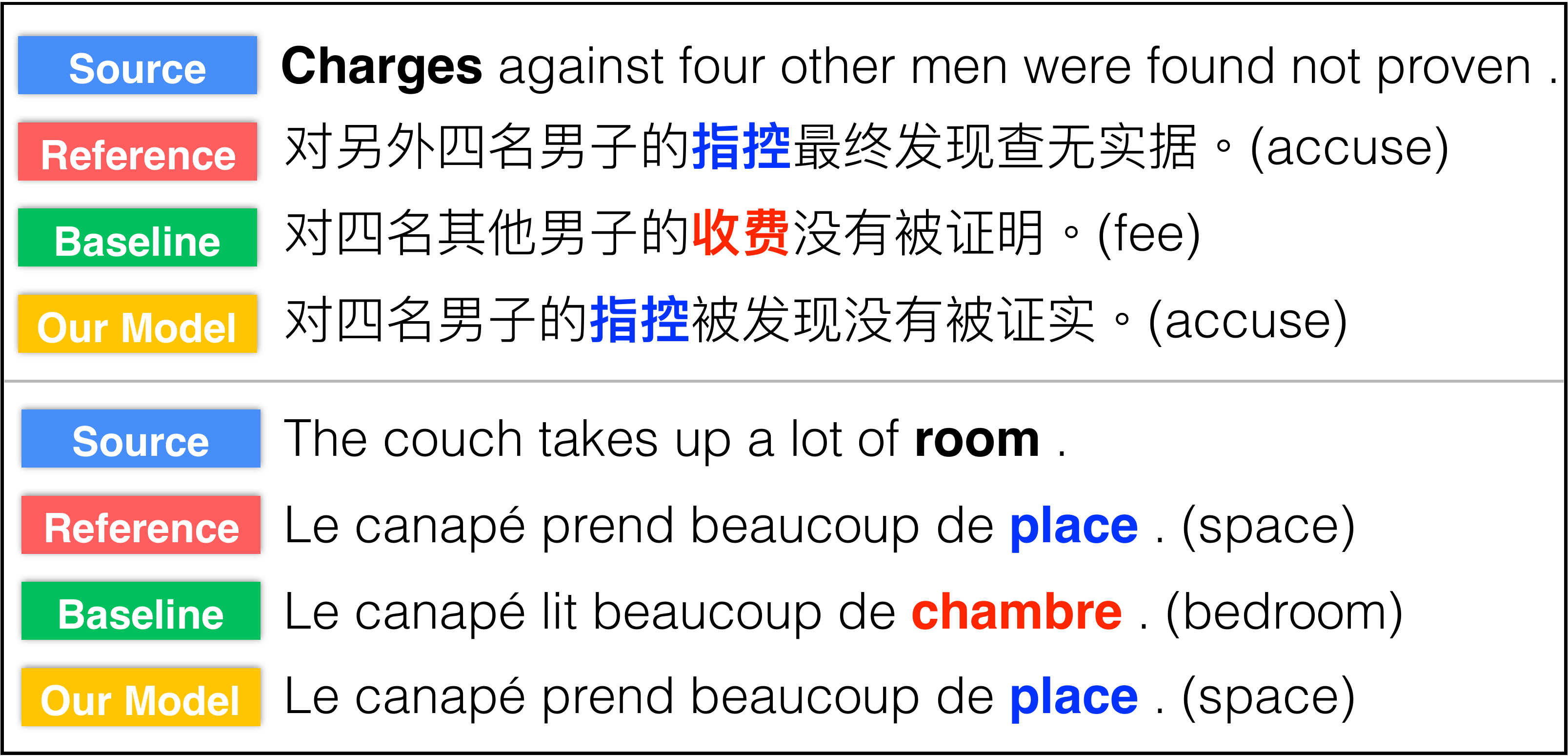}
\caption{Homographs where the baseline system makes mistakes (red words) but our proposed system incorporating a more direct representation of context achieves the correct translation (blue words). Definitions of corresponding blue and red words are in parenthesis.}
\label{fail_example}
\end{figure}

One other phenomenon that was poorly handled by PBMT was homographs -- words that have the same surface form but multiple senses.
As a result, PBMT systems required specific separate modules to incorporate long-term context, performing word-sense \cite{carpuat2007wsd,pu2017sense} or phrase-sense \cite{carpuat2007phrase} disambiguation to improve their handling of these phenomena.
Thus, we may wonder: do NMT systems suffer from the same problems when translating homographs? Or are the recurrent nets applied in the encoding step, and the strong language model in the decoding step enough to alleviate all problems of word sense ambiguity?

In \S\ref{sec:problems} we first attempt to answer this question quantitatively by examining the word translation accuracy of a baseline NMT system as a function of the number of senses that each word has.
Results demonstrate that standard NMT systems make a significant number of errors on homographs, a few of which are shown in Fig.~\ref{fail_example}.

With this result in hand, we propose a method for more directly capturing  contextual information that may help disambiguate difficult-to-translate homographs.
Specifically, we learn from neural models for word sense disambiguation \cite{kalchbrenner2014convolutional, iyyer2015deep,kaageback2016word, yuan2016semi,vsuster2016bilingual}, examining three methods inspired by this literature (\S\ref{sec:wsd}).
In order to incorporate this information into NMT,  we examine two methods: gating the word-embeddings in the model (similarly to \newcite{choi2017context}), and concatenating the context-aware representation to the word embedding (\S\ref{sec:addingcontext}).

To evaluate the effectiveness of our method, we 
compare our context-aware models with a strong baseline \cite{luong2015effective} on the English-German, English-French, and English-Chinese WMT dataset.
We show that our proposed model outperforms the baseline in the overall BLEU score across three different language pairs. Quantitative analysis demonstrates that our model performs better on translating homographs. Lastly, we show sample translations of the baseline system and our proposed model.

\section{Neural Machine Translation}
\label{sec:nmt}

We follow the global-general-attention NMT architecture with input-feeding proposed by \newcite{luong2015effective}, which we will briefly summarize here. The neural network  models the conditional distribution over translations $Y = (y_1, y_2, \dots, y_m)$ given a sentence in source language $X = (x_1, x_2, \dots x_n)$ as $P(Y|X)$. A NMT system consists of an encoder that summarizes the source sentence $X$ as a vector representation $\boldsymbol{h}$, and a decoder that generates a target word at each time step conditioned on both $\boldsymbol{h}$ and previous words. The conditional distribution is optimized with cross-entropy loss at each decoder output.

The encoder is usually a uni-directional or bi-directional RNN that reads the input sentence word by word. In the more standard bi-directional case, before being read by the RNN unit, each word in $X$ is mapped to an embedding in continuous vector space by a function $f_e$. 
\begin{align}
	f_e(x_t) = \boldsymbol{M_e}^{\top}\cdot\boldsymbol{1}(x_t)
\end{align}
$\boldsymbol{M_e} \in \mathcal{R}^{|V_s|\times d}$ is a matrix that maps a one-hot representation of $x_t$, $\boldsymbol{1}(x_t)$ to a $d$-dimensional vector space, and $V_s$ is the source vocabulary. We call the word embedding computed this way Lookup embedding.
The word embeddings are then read by a bi-directional RNN 
\begin{align}	\overrightarrow{\boldsymbol{h}}_t = \overrightarrow{\text{RNN}}_e(\overrightarrow{\boldsymbol{h}}_{t-1}, f_e(x_t)) \label{eq:bilstmfwd} \\
	\overleftarrow{\boldsymbol{h}}_t = \overleftarrow{\text{RNN}}_e(\overleftarrow{\boldsymbol{h}}_{t+1}, f_e(x_t))
\label{eq:bilstmbwd} 
\end{align}
After being read by both RNNs we can compute the actual hidden state at step $t$, $\boldsymbol{h}_t = [\overrightarrow{\boldsymbol{h}}_t;\overleftarrow{\boldsymbol{h}}_t]$, and the encoder summarized representation  $\boldsymbol{h} = \boldsymbol{h}_n$. The recurrent units $\overrightarrow{\text{RNN}}_e$ and $\overleftarrow{\text{RNN}}_e$ are usually either LSTMs \cite{hochreiter1997long} or GRUs \cite{chung2014empirical}.

The decoder is a uni-directional RNN that decodes the $t$th target word conditioned on (1) previous decoder hidden state $\boldsymbol{g}_{t-1}$, (2) previous word $y_{t-1}$ , and (3) the weighted sum of encoder hidden states $\boldsymbol{a}_t$. The decoder maintains the $t$th hidden state $g_t$ as follows,
\begin{align}
	\boldsymbol{g}_t = \overrightarrow{\text{RNN}}_d(\boldsymbol{g}_{t-1}, f_d(y_{t-1}), \boldsymbol{a}_t)
\end{align}
 Again, $\overrightarrow{\text{RNN}}_d$ is either LSTM or GRU, and $f_d$ is a mapping function in target language space.

The general attention mechanism for computing the weighted encoder hidden states $\boldsymbol{a}_t$ first computes the similarity between $\boldsymbol{g}_{t-1}$ and $\boldsymbol{h}_{t'}$ for $t' = 1,2,\dots,n$. 
\begin{align}
	\text{score}(\boldsymbol{g}_{t-1}, \boldsymbol{h}_{t'}) = \boldsymbol{g}_{t-1}\boldsymbol{W}_{att}\boldsymbol{h}^{\top}_{t'}
\end{align}
The similarities are then normalized through a softmax layer , which results in the weights for encoder hidden states.
\begin{align}
	\alpha_{t,t'} = \frac{\exp(\text{score}(\boldsymbol{g}_{t-1}, \boldsymbol{h}_{t'}))}{\sum_{k=1}^n \exp(\text{score}(\boldsymbol{g}_{t-1}, \boldsymbol{h}_{k}))}
\end{align}
We can then compute $\boldsymbol{a}_t$ as follows, 
\begin{align}
	\boldsymbol{a}_t = \sum_{k=1}^n \alpha_{t, k}\boldsymbol{h}_k
\end{align}
Finally, we compute the distribution over $y_t$ as,
\begin{align}
\hat{\boldsymbol{g}}_t = \text{tanh}(\boldsymbol{W}_1[\boldsymbol{g}_t;\boldsymbol{a}_t])\\
p(y_t|y_{<t}, X) = \text{softmax}(\boldsymbol{W}_2 \hat{\boldsymbol{g}}_t)
\end{align}

\begin{figure}[!t]
\centering
\includegraphics[width= 0.8\linewidth]{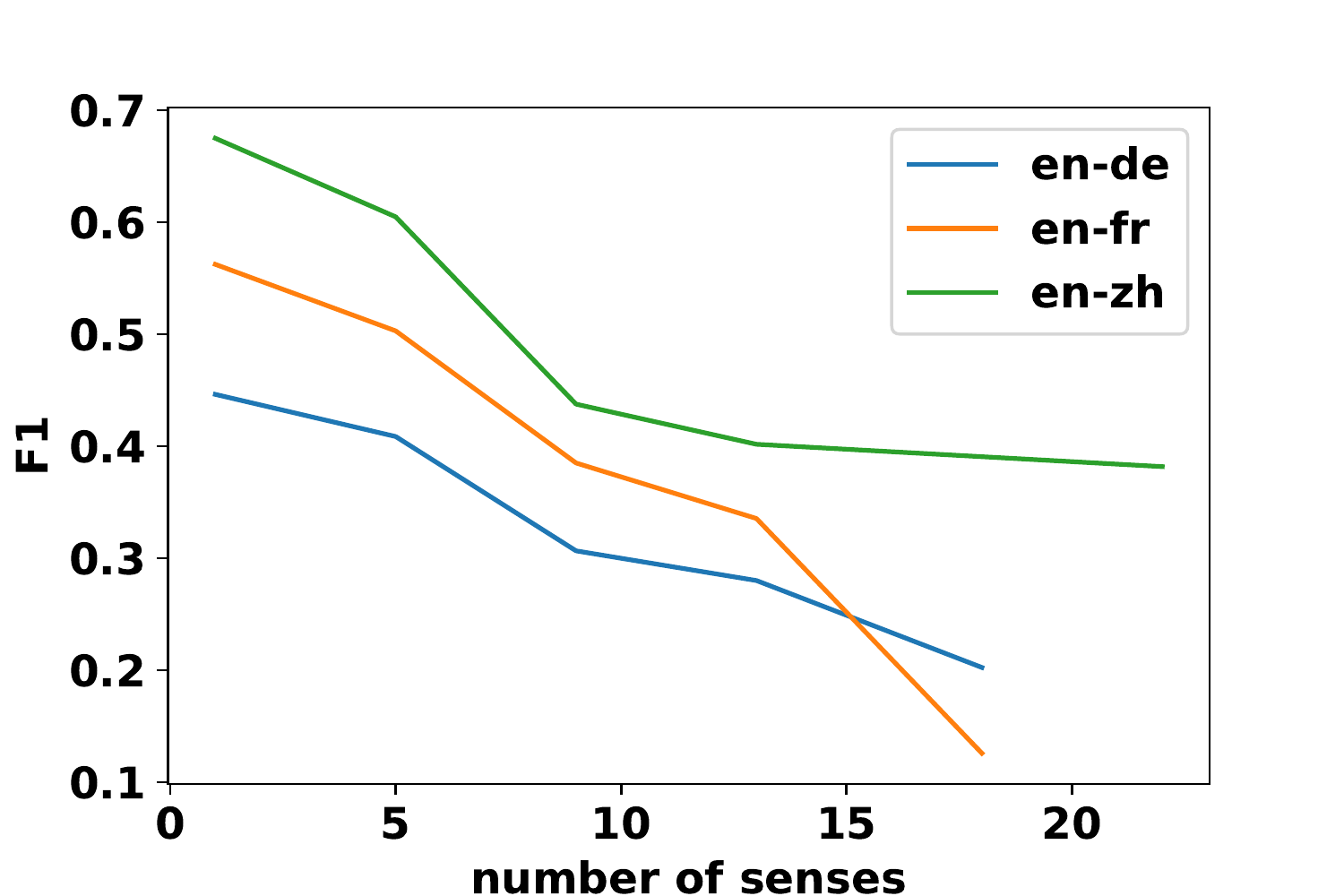}
\caption{Translation performance of words with different numbers of senses.}
\label{F1toRank}
\end{figure}

\section{NMT's Problems with Homographs}
\label{sec:problems}

As described in Eqs.~\eqref{eq:bilstmfwd} and~\eqref{eq:bilstmbwd}, NMT models encode the  words using recurrent encoders, theoretically endowing them with the ability to handle homographs through global sentential context.
However, despite the fact that they have this ability, our qualitative observation of NMT results revealed a significant number of ambiguous words being translated incorrectly, casting doubt on whether the standard NMT setup is able to appropriately learn parameters that disambiguate these word choices.

To demonstrate this more concretely, in Fig. ~\ref{F1toRank} we show the translation accuracy of an NMT system with respect to words of varying levels of ambiguity.
Specifically, we use the best baseline NMT system to translate three different language pairs from WMT test set (detailed in \S\ref{sec:exp}) and plot the F1-score of word translations by the number of senses that they have.
The number of senses for a word is acquired from the Cambridge English dictionary,\footnote{\url{http://dictionary.cambridge.org/us/dictionary/english/}} after excluding stop words.%
\footnote{We use the stop word list from NLTK \cite{bird2009natural}.}

We evaluate the translation performance of words in the source side by aligning them to the target side using \texttt{fast-align} \cite{dyer2013simple}.
The aligner outputs a set of target words to which the source words aligns for both the reference translation and the model translations. F1 score is calculated between the two sets of words.

After acquiring the F1 score for each word, we bucket the F1 scores by the number of senses, and plot the average score of four consecutive buckets as shown in Fig.~\ref{F1toRank}. As we can see from the results, the F1 score for words decreases as the number of senses increases for three different language pairs. This demonstrates that the translation performance of current NMT systems on words with more senses is significantly decreased from that for words with fewer senses.
From this result, it is  evident that modern NMT architectures are not enough to resolve the problem of homographs on their own.
The result corresponds to the findings in prior work \cite{rios2017improving}.

\section{Neural Word Sense Disambiguation}
\label{sec:wsd}

Word sense disambiguation (WSD) is the task of resolving the ambiguity of homographs \cite{ng1996integrating, mihalcea2004senselearner, zhong2010makes,di2013clustering,chen2014unified,camacho2015unified}, and  we hypothesize that by learning from these models we can improve the ability of the NMT model to choose the correct translation for these ambiguous words.
Recent research tackles this problem with neural models and has shown state-of-the art results on WSD datasets \cite{kaageback2016word, yuan2016semi}. In this section, we will summarize three methods for WSD which we will further utilize as three different \textit{context networks} to improve NMT. 

\paragraph{Neural bag-of-words (NBOW)}
\newcite{kalchbrenner2014convolutional, iyyer2015deep} have shown success by representing full sentences with a context vector, which is the average of the Lookup embeddings of the input sequence 
\begin{align}
	\boldsymbol{c}_t = \frac{1}{n}\sum_{k=1}^n \boldsymbol{M}_c^{\top}\boldsymbol{1}(x_k)
\end{align}
This is a simple way to model sentences, but has the potential to capture the global topic of the sentence in a straightforward and coherent way. However, in this case, the context vector would be the same for every word in the input sequence.

\paragraph{Bi-directional LSTM (BiLSTM)}
\newcite{kaageback2016word} leveraged a bi-directional LSTM that learns a context vector for the target word in the input sequence and predicts the word sense with a multi-layer perceptron. Specifically, we can compute the context vector $c_t$ for $t$th word similarly to bi-directional encoder as follows,
\begin{align}
	\overrightarrow{\boldsymbol{c}}_t = \overrightarrow{\text{RNN}}_c(\overrightarrow{\boldsymbol{c}}_{t-1}, f_c(x_t))
\end{align}
\begin{align}
	\overleftarrow{\boldsymbol{c}}_t = \overleftarrow{\text{RNN}}_c(\overleftarrow{\boldsymbol{c}}_{t+1}, f_c(x_t))
\end{align}
\begin{align}
	\label{eq:bilstmfwdbkwd}
	\boldsymbol{c}_t = [\overrightarrow{\boldsymbol{c}}_t;\overleftarrow{\boldsymbol{c}}_t]
\end{align}

$\overrightarrow{\text{RNN}}_c$, $\overleftarrow{\text{RNN}}_c$ are forward and backward LSTMs repectively, and $f_c(x_t) = \boldsymbol{M}_c^{\top}\boldsymbol{1}(x_t)$ is a function that maps a word to continous embedding space.

\paragraph{Held-out LSTM (HoLSTM)}
\newcite{yuan2016semi} trained a LSTM language model, which predicts a held-out word given the surrounding context, with a large amount of unlabeled text as training data. Given the context vector from this language model, they predict the word sense with a WSD classifier. Specifically, we can compute the context  vector $c_t$ for $t$th word by first replacing $t$th word with a special symbol (e.g. $<$\$$>$). We then feed the replaced sequence to a uni-directional LSTM:

\begin{align}
	\tilde{\boldsymbol{c}}_i = \overrightarrow{\text{RNN}}_c(\tilde{\boldsymbol{c}}_{i-1}, f_c(x_i))
\end{align}
Finally, we can get context vector for the $t$th word
\begin{align}
	\boldsymbol{c}_t = \tilde{\boldsymbol{c}}_n
\end{align}
$\overrightarrow{\text{RNN}}_c$ and $f_c$ are defined in BiLSTM paragraph, and $n$ is the length of the sequence. Despite the fact that the context vector is always the last hidden state of the LSTM no matter which word we are targeting, the input sequence read by the HoLSTM is actually different every time. 

\section{Adding Context to NMT}
\label{sec:addingcontext}

\begin{figure}[!t]
\centering
\includegraphics[height= 0.8\linewidth]{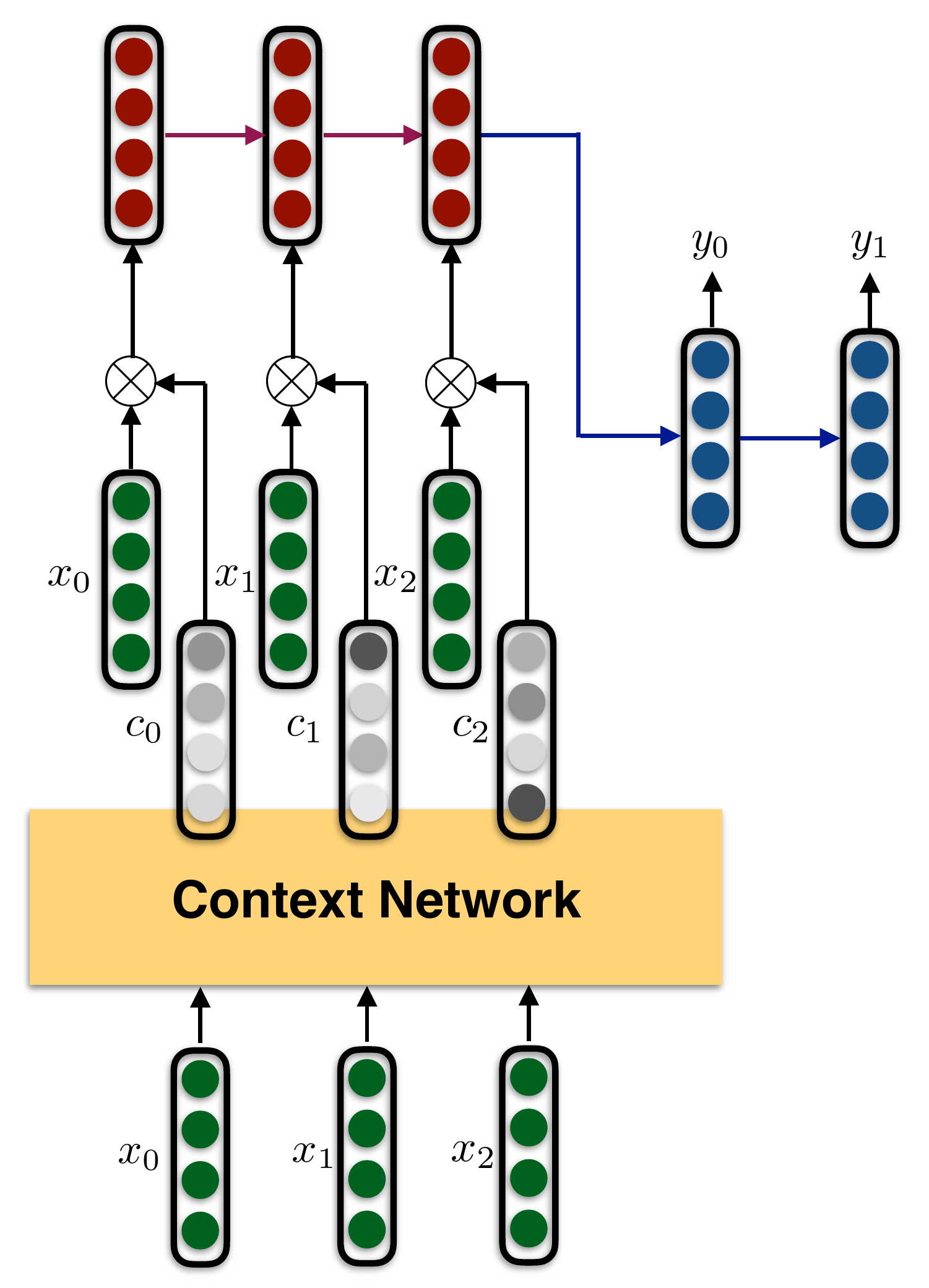}
\caption{Illustration of our proposed model. The context network is a differentiable network that computes context vector $c_t$ for word $x_t$ taking the whole sequence as input.  $\otimes$ represents the operation that combines original word embedding $x_t$ with corresponding context vector $c_t$ to form context-aware word embeddings. }
\label{model_concat}
\end{figure}

Now that we have several methods to incorporate global context regarding a single word, it is necessary to incorporate this context with NMT.
Specifically, we propose two methods to either \textit{Gate} or \textit{Concatenate} a context vector $\boldsymbol{c}_t$ with the Lookup embedding $\boldsymbol{M}_e^{\top}\cdot\boldsymbol{1}(x_t)$ to form a context-aware word embedding before feeding it into the encoder as shown in Fig.~\ref{model_concat}. The detail of these methods is described below. 

\paragraph{Gate}
Inspired by \newcite{choi2017context}, as our first method for integration of context-aware word embeddings, we use a gating function as follows:
\begin{align}
	f'_e(x_t) 
    &= f_e(x_t)\odot\sigma(\boldsymbol{c}_t)\\
    &= \boldsymbol{M}_e^{\top}\boldsymbol{1}(x_t)\odot\sigma(\boldsymbol{c}_t)
\end{align}
The symbol $\odot$ represents element-wise multiplication, and $\sigma$ is element-wise sigmoid function. 
\newcite{choi2017context} use this method in concert with averaged embeddings from words in source language like the NBOW model above, which naturally uses the same context vectors for all time steps.
In this paper, we additionally test this function with context vectors calculated using the BiLSTM and HoLSTM .

\paragraph{Concatenate}
We also propose another way for incorporating context: by concatenating the context vector with the word embeddings.
This is expressed as below:
\begin{align}
	\label{eq:concatenate}
	f'_e(x_t) 
    &= \boldsymbol{W}_3[f_e(x_t);\boldsymbol{c}_t]\\
    &= \boldsymbol{W}_3[\boldsymbol{M}_e^{\top}\boldsymbol{1}(x_t);\boldsymbol{c}_t]
\end{align}
$\boldsymbol{W}_3$ is used to project the concatenated vector back to the original $d$-dimensional space.

For each method can compute context vector $\boldsymbol{c}_t$ with either the NBOW, BiLSTM, or HoLSTM described in \S\ref{sec:wsd}. We share the parameters in $f_e$ with $f_c$ (i.e. $\boldsymbol{M}_e = \boldsymbol{M}_c$) since the vocabulary space is the same for context network and encoder. As a result, our context network only slightly increases the number of model parameters. Details about the number of parameters of each model we use in the experiments are shown in Table~\ref{baseline_compare}.

\begin{table*}[t!]
\centering
\resizebox{0.9\textwidth}{!}
{\begin{tabular}{cc|ccc|c|cc}
\textbf{Context}  & \multicolumn{1}{c}{\textbf{Integration}}         &   \multicolumn{1}{c}{\textbf{uni/bi}}   & \multicolumn{1}{c}{\textbf{\#layers}}    & \multicolumn{1}{c}{\textbf{\#params}}    & \multicolumn{1}{c}{\textbf{Ppl}}              & \multicolumn{1}{c}{\textbf{WMT14}}              & \textbf{WMT15}        \\ \hline\hline
None & \multicolumn{1}{c|}{-}                           & $\rightarrow$               & 2                    & 85M & 7.12 & 20.49 & 22.95                \\
None & \multicolumn{1}{c|}{-}                           & $\leftrightarrow$                   & 2                    & 83M & 7.20  & \textbf{21.05} & \textbf{23.83}                \\
None & \multicolumn{1}{c|}{-}                           & $\leftrightarrow$                   & 3                    & 86M & 7.50  & 20.86 & 23.14                \\
\hline\hline
NBOW & Concat & $\rightarrow$               & 2                    & 85M & 7.23 & 20.44 & 22.83                \\
NBOW & Concat & $\leftrightarrow$                   & 2                    & 83M & 7.28 & 20.76 & 23.61                \\
HoLSTM & Concat & $\rightarrow$               & 2                    & 87M & 7.19 & 20.67 & 23.05                \\
HoLSTM & Concat& $\leftrightarrow$                   & 2                    & 86M & 7.04     & 21.15      &     23.53                 \\
BiLSTM & Concat & $\rightarrow$               & 2                    & 87M & 6.88 & \textbf{21.80}  & \textbf{24.52}                \\
BiLSTM & Concat & $\leftrightarrow$                   & 2                    & 85M & 6.87 & 21.33 & 24.37  \\ \hline\hline

NBOW & Gating & $\rightarrow$               & 2                    & 85M & 7.14     & 20.20      &  22.94                    \\

NBOW & Gating & $\leftrightarrow$                   & 2                    & 83M & 6.92     & 21.16      &    23.52   \\  
BiLSTM & Gating & $\rightarrow$               & 2                    & 87M & 7.07     & 20.94      &  23.58                    \\
BiLSTM & Gating & $\leftrightarrow$                   & 2                    & 85M & 7.11     & 21.33      &    24.05              
\end{tabular}}
\caption{\textbf{WMT'14, WMT'15 English-German results} - We show perplexities (Ppl) on development set and tokenized BLEU on WMT'14 and WMT'15 test set of various NMT systems. We also show different settings for different systems. $\rightarrow$ represents uni-directional, and $\leftrightarrow$ represents bi-directional. We also highlight the best baseline model and the best proposed model in bold. The best baseline model will be referred as \textit{base} or \textit{baseline} and the best proposed model will referred to as \textit{best} for further experiments.}
\label{baseline_compare}
\end{table*}
\section{Experiments}
\label{sec:exp}
We evaluate our model on three different language pairs:
	English-French (WMT'14), and English-German (WMT'15), English-Chinese (WMT'17) with English as the source side. For German and French, we use a combination of Europarl v7, Common Crawl, and News Commentary as training set. For development set, newstest2013 is used for German and newstest2012 is used for French. For Chinese, we use a combination of News Commentary v12 and the CWMT Corpus as the training set and held out 2357 sentences as the development set. Translation
performances are reported in case-sensitive
BLEU on newstest2014 (2737 sentences), newstest2015 (2169 sentences) for German, newstest2013 (3000 sentences), newstest2014 (3003 sentences) for French, and newsdev2017 (2002 sentences) for Chinese.\footnote{We use the development set as testing data because the official test set hasn't been released.} Details about tokenization are as follows. For German, we use the tokenized dataset from \newcite{luong2015effective}; for French, we used the moses \cite{koehn2007moses} tokenization script with the ``-a'' flag; for Chinese,  we split sequences of Chinese characters, but keep sequences of non-Chinese characters as they are, using the script from IWSLT Evaluation 2015.%
\footnote{\url{https://sites.google.com/site/iwsltevaluation2015/mt-track}}

We compare our context-aware NMT systems with strong baseline models on each dataset.

\begin{table}[t!]
\centering
\begin{tabular}{c|cc}
System                      & \multicolumn{2}{c}{BLEU}                               \\ \hline
\textbf{en $\rightarrow$ de} & \multicolumn{1}{c|}{\textbf{WMT'14}} & \textbf{WMT'15} \\ \hline
baseline                & \multicolumn{1}{c|}{21.05}           & 23.83           \\
best                       & \multicolumn{1}{c|}{\textbf{21.80}}            & \textbf{24.52}           \\ \hline
\textbf{en $\rightarrow$ fr} & \multicolumn{1}{c|}{\textbf{WMT'13}} & \textbf{WMT'14} \\ \hline
baseline                & \multicolumn{1}{c|}{28.21}           & 31.55           \\ 
best                       & \multicolumn{1}{c|}{\textbf{28.77}}           & \textbf{32.39}           \\ \hline
\textbf{en $\rightarrow$ zh} & \multicolumn{2}{c}{\textbf{WMT'17}}                   \\ \hline
baseline                & \multicolumn{2}{c}{24.07}                              \\ 
best                       & \multicolumn{2}{c}{\textbf{24.81}}                              
\end{tabular}
\caption{\textbf{Results on three different language pairs} - The best proposed models (BiLSTM+Concat+uni) are significantly better (p-value $ < 0.001$) than baseline models using paired bootstrap resampling \cite{koehn2004statistical}. }
\label{multiple_languages}
\end{table}

\subsection{Training Details}
We limit our vocabularies to
be the top 50K most frequent words for both source and target language.
Words not in these shortlisted vocabularies
are converted into an $\langle$unk$\rangle$ token.

When training our NMT systems, following
\newcite{bahdanau2015align}, we filter
out sentence pairs whose lengths exceed
50 words and shuffle mini-batches as we proceed. We train our model with the following settings using SGD as our optimization method. (1)  We start with a learning rate of 1 and we begin to halve the learning
rate every epoch once it overfits. \footnote{We define overfitting to be when perplexity on the dev set of the current epoch is worse than the previous epoch.} (2) We train until the model converges. (i.e. the difference between the perplexity for the current epoch and the previous epoch is less than 0.01) (3) We batched the instances with the same length and our maximum mini-batch size is 256,
and (4) the normalized gradient is rescaled whenever its norm exceeds 5. (6) Dropout is applied between vertical RNN stacks with probability 0.3. Additionally, the context network is trained jointly with the encoder-decoder architecture.
Our model is built upon OpenNMT ~\cite{2017opennmt} with the default settings unless otherwise noted.

\begin{table*}[t!]
\centering
\resizebox{\textwidth}{!}{
\begin{tabular}{c|c|c|c|c|ccc}
\multirow{2}{*}{language}            & \multirow{2}{*}{System} & \multicolumn{3}{c|}{Homograph}                   & \multicolumn{3}{c}{All Words}                                                                         \\ \cline{3-8} 
                                     &                         & F1             & Precision      & Recall         & \multicolumn{1}{c|}{F1}             & \multicolumn{1}{c|}{Precision}      & Recall                    \\ \hline
\multirow{2}{*}{en $\rightarrow$ de}  & baseline                & 0.401          & 0.422          & 0.382          & \multicolumn{1}{c|}{0.547}          & \multicolumn{1}{c|}{0.569}          & 0.526                     \\
                                     & best                    & 0.426 (+\textit{0.025}) & 0.449 (+\textit{0.027}) & 0.405 (+\textit{0.023}) & \multicolumn{1}{c|}{0.553 (+\textit{0.006})} & \multicolumn{1}{c|}{0.576 (+\textit{0.007})} & 0.532 (+\textit{0.006}) \\ \hline
\multirow{2}{*}{en $\rightarrow$ fr} & baseline                & 0.467          & 0.484          & 0.451          & \multicolumn{1}{c|}{0.605}          & \multicolumn{1}{c|}{0.623}          & 0.587                     \\
                                     & best                    & 0.480 (+\textit{0.013}) & 0.496 (+\textit{0.012}) & 0.465 (+\textit{0.014}) & \multicolumn{1}{c|}{0.613 (+\textit{0.008})} & \multicolumn{1}{c|}{0.630 (+\textit{0.007})} & 0.596 (+\textit{0.009}) \\ 
\hline
\multirow{2}{*}{en $\rightarrow$ zh} & baseline                & 0.578          & 0.587          & 0.570          & \multicolumn{1}{c|}{0.573}          & \multicolumn{1}{c|}{0.605}          & 0.544                     \\
                                     & best                    & 0.590 (+\textit{0.012}) & 0.599 (+\textit{0.012}) & 0.581 (+\textit{0.011}) & \multicolumn{1}{c|}{0.581 (+\textit{0.008})} & \multicolumn{1}{c|}{0.612 (+\textit{0.007})} & 0.552 (+\textit{0.008})
\end{tabular}}
\caption{Translation results for homographs and all words in our NMT vocabulary. We compare scores for baseline and our best proposed model on three different language pairs. Improvements are in italic. We performed bootstrap resampling for 1000 times: our best model improved more on homographs than all words in terms of either f1, precision, or recall with $p<0.05$, indicating statistical significance across all measures.}
\label{fast_align}
\end{table*}
\subsection{Experimental Results}
\label{sec:compare_baseline}
In this section, we compare our proposed context-aware NMT models with baseline models on English-German dataset. Our baseline models are encoder-decoder models using global-general attention and input feeding on the decoder side as described in \S\ref{sec:nmt}, varying the settings on the encoder side. Our proposed model builds upon baseline models by concatenating or gating different types of context vectors. We use LSTM for encoder, decoder, and context network. The decoder is the same across baseline models and proposed models, having 500 hidden units. During testing, we use beam search with a beam size of 5. The dimension for input word embedding $d$ is set to 500 across encoder, decoder, and context network. Settings for three different baselines are listed below. 

\begin{description}
\item[Baseline 1:]
An uni-directional LSTM with 500 hidden units and 2 layers of stacking LSTM.
\item[Baseline 2:]
A bi-directional LSTM with 250 hidden units and 2 layers of stacking LSTM. Each state is summarized by concatenating the hidden states of forward and backward encoder into 500 hidden units.
\item[Baseline 3:]
A bi-directional LSTM with 250 hidden units and 3 layers of stacking LSTM.
This can be compared with the proposed method, which adds an extra layer of computation before the word embeddings, essentially adding an extra layer.
\end{description}
The context network uses the below settings.
\begin{description}
\item[NBOW:]
Average word embedding of the input sequence.
\item[BiLSTM:]
A single-layer bi-directional LSTM with 250 hidden units. The context vector is represented by concatenating the hidden states of forward and backward LSTM into a 500 dimensional vector.
\item[HoLSTM:]
A single-layer uni-directional LSTM with 500 hidden units.
\end{description}

The results are shown in Table ~\ref{baseline_compare}.
The first thing we observe is that the best context-aware model (results in bold in the table) achieved improvements of around 0.7 BLEU on both WMT14 and WMT15 over the respective baseline methods with 2 layers.
This is in contrast to simply using a 3-layer network, which actually degrades performance, perhaps due to the vanishing gradients problem it increases the difficulty in learning.

Next, comparing different methods for incorporating context, we can see that BiLSTM performs best across all settings. HoLSTM performs slightly better than NBOW, and NBOW obviously suffers from having the same context vector for every word in the input sequence failing to outperform the corresponding baselines. 
Comparing the two integration methods that incorporate context into word embeddings. Both methods improve over the baseline with BiLSTM as the context network. Concatenating the context vector and the word embedding performed better than gating. 
Finally, in contrast to the baseline, it is not obvious whether using uni-directional or bi-directional as the encoder is better for our proposed models, particularly when BiLSTM is used for calculating the context network.
This is likely due to the fact that bi-directional information is already captured by the context network, and may not be necessary in the encoder itself.

We further compared the two systems on two different languages, French and Chinese. We achieved 0.5-0.8 BLEU improvement, showing our proposed models are stable and consistent across different language pairs. The results are shown in Table ~\ref{multiple_languages}. 

To show that our 3-layer models are properly trained, we ran a 3-layer bidirectional encoder with residual networks on En-Fr and got 27.45 for WMT13 and 30.60 for WMT14, which is similarly lower than the two layer result. It should be noted that previous work such as ~\newcite{britz2017massive} have also noted that the gains for encoders beyond two layers is minimal.

\subsection{Targeted Analysis} 
In order to examine whether our proposed model can better translate words with multiple senses, we evaluate our context-aware model on a list of homographs extracted from Wikipedia\footnote{ \url{https://en.wikipedia.org/wiki/List_of_English_homographs}} compared to the baseline model on three different language pairs. For the baseline model, we choose the best-performing model, as described in \S\ref{sec:compare_baseline}.

To do so, we first acquire the translation of homographs in the source language using \texttt{fast-align} \cite{dyer2013simple}. We run \texttt{fast-align} on all the parallel corpora including training data and testing data\footnote{Reference translation, and all the system generated translations.} because the unsupervised nature of the algorithm requires it to have a large amount of training data to obtain accurate alignments. The settings follow the default command on fast-align github page including heuristics combining forward and backward alignment. 
Since there might be multiple aligned words in  the target language given a word in source language, we treat a match between the aligned translation of a targeted word of the reference and the translation of a given model as true positives and use F1, precision, and recall as our metrics, and take the micro-average across all the sentence pairs.
\footnote{The link to the evaluation script -- \\ \url{https://goo.gl/oHYR8E}}
We calculated the scores for the 50000 words/characters from our source vocabulary using only English words.
The results are shown in Table ~\ref{fast_align}.
The table shows two interesting results:
(1) The score for the homographs is lower than the score obtained from all the words in the vocabulary.
This shows that words with more meanings are harder to translate with Chinese as the only exception.%
\footnote{One potential explanation for Chinese is that because the Chinese results are generated on the character level, the automatic alignment process was less accurate.}
(2) The improvement of our proposed model over baseline model is larger on the homographs compared to all the words in vocabulary. This shows that although our context-aware model is better overall, the improvements are particularly focused on words with multiple senses, which matches the intuition behind the design of the model.

\subsection{Qualitative Analysis}
\begin{CJK*}{UTF8}{gbsn}
We show sample translations on English-Chinese WMT'17 dataset in Table~\ref{samples} with  three kinds of examples. We highlighted the English homograph in bold, correctly translated words in blue, and wrongly translated words in red. (1) Target homographs are translated into the correct sense with the help of context network. For the first sample translation, ``meets'' is correctly translated to ``会见'' by our model, and  wrongly translated to  ``符合'' by baseline model.  In fact, ``会见'' is closer to the definition ``come together intentionally'' and ``符合'' is closer to "satisfy" in the English dictionary. (2) Target homographs are translated into different but similar senses for both models in the forth example. Both models translate the word ``believed'' to common translations ``被认为'' or
``相信'', but these meaning are both close to reference translation ``据信''.  (3) Target homograph is translated into the wrong sense for the baseline model, but is not translated in our model in the fifth example.
\end{CJK*}
\begin{table*}[t!]
\begin{CJK*}{UTF8}{gbsn}
\centering
\resizebox{\textwidth}{!}{\begin{tabular}{cl}
\hline
\multicolumn{2}{l}{English-Chinese Translations}                                                                                                                            \\ \hline
\multicolumn{1}{l|}{src}  & Ugandan president \textbf{meets} Chinese FM , anticipates closer cooperation                                               \\
\multicolumn{1}{l|}{ref}  & 乌 干 达 总 统 \textcolor{blue}{\textbf{会 见}} 中 国 外 长 ， 期 待 增 进 合 作 (come together intentionally)                               \\
\multicolumn{1}{l|}{best} & 乌 干 达 总 统 \textcolor{blue}{\textbf{会 见}} 中 国 调 频 ， 预 期 更 密 切 合 作 (come together intentionally)                             \\
\multicolumn{1}{l|}{base} & 乌 干 达 总 统 \textcolor{red}{\textbf{符 合} } 中 国 调 频 ， 预 期 更 加 合 作 (satisfy)                                                   \\ \hline
\multicolumn{1}{l|}{src}  & Investigators are trying to \textbf{establish} whether Kermiche and Petitjean had accomplices in France and                \\
\multicolumn{1}{l|}{}     & whether they had links with Islamic State , which has claimed responsibility for the attack .                                                   \\
\multicolumn{1}{l|}{ref}  & 调 查 人 员 正 试 图\textcolor{blue}{\textbf{ 确 定}} 克 尔 米 奇 和 帕 迪 让 在 法 国 是 否 有 同 谋 ， 以 及 是 否 与 伊 斯 兰 国 武 装 分 子 有 联 系，            \\
\multicolumn{1}{l|}{}     & 伊 斯 兰 国 武 装 分 子 声 称 对 此 次 袭 击 负 责 。 (get proof of something)                                                                                    \\
\multicolumn{1}{l|}{best} & 调 查 人 员 正 试 图\textcolor{blue}{\textbf{ 确 定}} Kermiche 和 Petitjean 在 法 国 是 否 有 同 谋 ， 他 们 是 否 与 伊 斯 兰 国 有 联 系 ，              \\
\multicolumn{1}{l|}{}     & 声 称 对 这次 袭 击 负 责 。 (get proof of something)                                                                                                     \\
\multicolumn{1}{c|}{base} & 调 查 人 员 正 在 努 力\textcolor{red}{\textbf{ 建 立}} 法 国 的 同 谋 和 他 们 是 否 与 伊 斯 兰 国 有 联 系 , 该 国 声 称 对 这 次 袭 击 负 有 责 任 。 (to start) \\ \hline
\multicolumn{1}{l|}{src}  & The decrease of transaction settlement fund \textbf{balance} in the securities market in July was smaller than that in June ,                   \\
\multicolumn{1}{l|}{}     & while the net bank @-@ securities transfers stood at negative RMB 66.6 billion .                                                                \\
\multicolumn{1}{l|}{ref}  & 7 月 证 券 市 场 交 易 结 算 资 金 \textcolor{blue}{\textbf{余 额}} 减 少 额 较 6 月 大 幅 降 低 ， 银 证 转 账 变 动 净 额 为 - 666 亿 元 。 (money left)                         \\
\multicolumn{1}{l|}{best} & 7 月 份 证 券 市 场 交 易 结 算 资 金 \textcolor{blue}{\textbf{余 额}} 的 减 少 小 于 6 月 份 ， 而 银 行 证 券 转 让 净 额 为 negative 亿 元 。 (money left)                      \\
\multicolumn{1}{l|}{base} & 七 月 证 券 市 场 交 易 结 算 基 金 \textcolor{red}{\textbf{平 衡}} 的 减 少 比 六 月 份 小 ， 而 净 银 行 证 券 转 让 则 为 负 元 。 (equal weight or force)                       \\ \hline
\hline
\multicolumn{1}{l|}{src}  & Initial reports suggest that the gunman may have shot a woman , \textbf{believed} to be his ex @-@ partner .                                    \\
\multicolumn{1}{l|}{ref}  & 据 初 步 报 告 显 示 ， 开 枪 者 可 能 击 中 一 名 妇 女 ， \textcolor{blue}{\textbf{据 信}} 是 他 的 前 搭 档 。 (been accepted as truth)                                                            \\
\multicolumn{1}{l|}{best} & 初 步 的 报 道 表 明 ， 枪 手 可 能 已 经 射 杀 了 一 个 女 人 ， \textcolor{red}{\textbf{被 认 为}} 是 他 的 前 伙 伴 。 (been known as)                                                      \\
\multicolumn{1}{l|}{base} & 最 初 的 报 道 显 示 ， 枪 手 可 能 已 经 射 杀 了 一 名 妇 女 ， \textcolor{red}{\textbf{相 信}} 他 是 他 的 前 伙 伴 。 (accept as truth)                                                      \\ \hline\hline
\multicolumn{1}{l|}{src}  & When the game came to the last 3 ' 49 ' ' , Nigeria \textbf{closed} to 79 @-@ 81 after Aminu added a layup .                                    \\
\multicolumn{1}{l|}{ref}  & 比 赛 还 有 3 分 49 秒 时 ， 阿 米 努 上 篮 得 手 后 ， 尼 日 利 亚 将 比 分 \textcolor{blue}{\textbf{追 成}} 了 79-81 。 (narrow)                                                  \\
\multicolumn{1}{l|}{best} & 当 这 场 比 赛 到 了 最 后 三 个 “ 49 ” 时 ， 尼 日 利 亚 在 Aminu 增 加 了 一 个 layup 之 后 \underline{\textcolor{red}{\textbf{MISSING TRANSLATION}}}。                        \\
\multicolumn{1}{l|}{base} & 当 游 戏 到 达 最 后 3 “ 49 ” 时 ， 尼 日 利 亚 已 经 \textcolor{red}{\textbf{关 闭}} 了 Aminu 。 (end)                                                                 \\ \hline

\end{tabular}}
\caption{\textbf{Sample translations} - for each example, we show sentence in source language (src), the human translated reference (ref), the translation generated by our best context-aware model (best), and the translation generated by baseline model (base). We also highlight the word with multiple senses in source language in bold, the corresponding correctly translated words in blue and wrongly translated words in red. The definitions of words in blue or red are in parenthesis.}
\label{samples}
\end{CJK*}
\end{table*}

\section{Related Work}
Word sense disambiguation (WSD), the task of determining the
correct meaning or sense of a word in context is a long standing task in NLP ~\cite{yarowsky1995unsupervised, ng1996integrating, mihalcea2004senselearner, navigli2009word, zhong2010makes,di2013clustering,chen2014unified,camacho2015unified}. 
Recent research on tackling WSD and capturing multi-senses includes work leveraging LSTM ~\cite{kaageback2016word, yuan2016semi}, which we extended as a context network in our paper and predicting senses with word embeddings that capture context. \newcite{vsuster2016bilingual, kawakami2015learning} also showed that bilingual data improves WSD.
In contrast to the standard WSD formulation,
~\newcite{vickrey2005word} reformulated the task of WSD for Statistical Machine Translation (SMT) as predicting possible target translations which directly improves the accuracy of machine translation. Following this reformulation, \newcite{chan2007word, carpuat2007phrase,carpuat2007wsd} integrated WSD systems into phrase-based systems. \newcite{xiong2014sense} breaks the process into two stages. First predicts the sense of the ambiguous source word. The predicted
word senses together with other context
features are then used to predict possible
target translation. 
Within the framework of Neural MT, there are works that has similar motivation to ours. \newcite{choi2017context} leverage the NBOW as context and gate the word-embedding on both encoder and decoder side.
However, their work does not distinguish context vectors for words in the same sequence, in contrast to the method in this paper, and our results demonstrate that this is an important feature of methods that handle homographs in NMT. In addition, our quantitative analysis of the problems that homographs pose to NMT and evaluation of how context-aware models fix them was not covered in this previous work. \newcite{rios2017improving} tackled the problem by adding sense embedding learned with additional corpus and evaluated the performance on the sentence level with contrastive translation.

\section{Conclusion}
Theoretically, NMT systems should be able to handle homographs if the encoder captures the clues to translate them correctly. In this paper, we empirically show that this may not be the case; the performance of word level translation degrades as the number of senses for each word increases. We hypothesize that this is due to the fact that each word is mapped to a word vector despite them being in different contexts, and propose to integrate methods from neural WSD systems into an NMT system to alleviate this problem. We concatenated the context vector computed from the context network with the word embedding to form a context-aware word embedding, successfully improving the NMT system. We evaluated our model on three different language pairs and outperformed a strong baseline model according to BLEU score in all of them. We further evaluated our results targeting the translation of homographs, and our model performed better in terms of F1 score.

While the architectures proposed in this work do not ~\textit{solve} the problem of homographs, our empirical results in Table~\ref{fast_align} demonstrate that they do yield improvements (larger than those on other varieties of words).
We hope that this paper will spark discussion on the topic, and future work will propose even more focused architectures.
\bibliography{naaclhlt2018}
\bibliographystyle{acl_natbib}

\end{document}